\documentclass[table,xcdraw]{article} 
\usepackage{iclr2023_conference,times}

\usepackage{amsmath,amsfonts,bm}

\def\eqref#1{equation~\ref{#1}}

\def\1{\bm{1}}

\DeclareMathAlphabet{\mathsfit}{\encodingdefault}{\sfdefault}{m}{sl}
\SetMathAlphabet{\mathsfit}{bold}{\encodingdefault}{\sfdefault}{bx}{n}

\def\gS{{\mathcal{S}}}

\DeclareMathOperator*{\argmin}{arg\,min}

\usepackage{hyperref}
\usepackage{url}

\newcommand{\norm}[1]{\left\|#1\right\|}

\usepackage[vlined,ruled]{algorithm2e}

\usepackage[utf8]{inputenc} 
\usepackage[T1]{fontenc}    
\usepackage{hyperref}       
\usepackage{url}            
\usepackage{booktabs}       
\usepackage{amsfonts}       
\usepackage{nicefrac}       
\usepackage{microtype}      
\usepackage{thm-restate}

\usepackage{microtype}
\usepackage{graphicx}
\usepackage{subfigure}
\usepackage{changes}

\usepackage{amsmath}
\usepackage{multirow}
\usepackage{hhline}
\usepackage{wrapfig}
\usepackage{floatrow}
\usepackage{caption}
\usepackage{enumitem}
\usepackage[export]{adjustbox}
\usepackage{amsmath,mathtools}

\usepackage{tikz}
\def\checkmark{\tikz\fill[scale=0.4](0,.35) -- (.25,0) -- (1,.7) -- (.25,.15) -- cycle;}

\usepackage{algorithmic}

\makeatletter
\newcommand{\raisemath}[1]{\mathpalette{\raisem@th{#1}}}
\newcommand{\raisem@th}[3]{\raisebox{#1}{$#2#3$}}
\makeatother

\newcommand{\uglad}{{\texttt{uGLAD}~}}
\newcommand{\glad}{{\texttt{GLAD}~}}

\title{Methods for recovering conditional independence graphs: A survey} 

\author{
  Harsh Shrivastava~~~~
  Urszula Chajewska
  \hspace{0mm}\\
  \hspace{-3mm}
  \begin{tabular}{c}
      $\prescript{}{}{\quad~~\text{Microsoft Research, Redmond, USA}}$
  \end{tabular}
}
\newcommand\independent{\protect\mathpalette{\protect\independenT}{\perp}}
\def\independenT#1#2{\mathrel{\rlap{$#1#2$}\mkern2mu{#1#2}}}

\newcommand{\Rho}{\mathrm{P}}

\iclrfinalcopy 
\begin{document}

\maketitle

\begin{abstract}
Conditional Independence (CI) graphs are a type of probabilistic graphical models that are primarily used to gain insights about feature relationships. Each edge represents the partial correlation between the connected features which gives information about their direct dependence.
In this survey, we list out different methods and study the advances in techniques developed to recover CI graphs. We cover traditional optimization methods as well as recently developed deep learning architectures along with their recommended implementations
. To facilitate wider adoption, we include preliminaries that consolidate associated operations, for example techniques to obtain covariance matrix for mixed datatypes. 

\end{abstract}

\section{Introduction}

Given an input data $X\in\mathbb{R}^{M\times D}$ with $M$ samples and $D$ features generated by an unknown distribution $P$. Let $\{X_1, \cdots, X_D\}$ represent the individual features. It is often beneficial to know which features are directly correlated to which other features. This can help us understand the input data better by giving a feature inter-dependence overview and also assist in taking system design decisions. It is important to understand the difference between two features $X_i$ and $X_j$ that are correlated vs being directly correlated. Consider a universe with only three variables as an undirected graph $G_{\text{ex}}$ = [study]---[grades]---[graduation], where we assume that each edge represents a positive correlation. We can see that if a student studies, then they will get good grades, which in turn increases their chances of graduation. We can thus conclude that study is correlated to graduation. But, if we know a student's grades, regardless of whether the student has studied, we can make conclusions about the chances of graduation. Thus, studying is not directly correlated to the graduation, whereas grades and graduation are directly correlated. Taking clue from this toy example, we can envision that such analysis can be very useful to obtain valuable insights from the input data as well as leverage the natural interpretability provided by the graph representation. 

Formally, the conditional independence (CI) graph of a set of random variables $X_i's$ is the undirected graph $G=(D,E)$ where $D=\{1,2,…,d\}$ nodes and $(i,j)$ is not in the edge set if and only if $X_i\independent X_j|\textbf{X}_{D\backslash{i,j}}$, where $\textbf{X}_{D\backslash{i,j}}$ denotes the collection of all of the random variables except for $X_i$ and $X_j$. 
We assume that the set of conditional independence properties encoded in the graph reflects independence properties of the distribution used to generate input data.  More formally, 
let's define $\mathcal{I}(P)$ to be the set of independence assertions of the form $(X_i\independent X_j | X_k)$ that hold in the data generating distribution $P$.  If $P$ satisfies all independence assertions encoded by $G$, that is, $\mathcal{I}(G) \subseteq \mathcal{I}(P)$, we say that $G$ is an I-map (independence map) of $P$.  Since the complete graph is an I-map for any distribution, we are typically interested in a minimal I-map, that is, an I-map graph such that a removal of a single edge will render it not an I-map. If  $\mathcal{I}(G) = \mathcal{I}(P)$, the graph $G$ is a perfect map of $P$.  However, it is important to keep in mind that not every distribution has a perfect map.

The conditional independence graph can be parameterized by learning edge weights that represent
partial correlations between the features.  In that case the edge weights will range between $e_w\sim(-1, 1)$. In the rest of the paper, we will use the term \textit{conditional independence graphs} to denote graphs with such a parameterization.

CI graphs are primarily used to gain insights about feature relationships to help with decision making. In some cases, they are also used to study the evolving feature relationships with time. The focus of this paper is to review
different methods and recent techniques developed to recover CI graphs.
We will start by giving a brief overview of algorithms that recover different types of graphs. 

\subsection{Umbrella of Graph Recovery Approaches}
\begin{figure}
\centering 
\includegraphics[width=135mm]{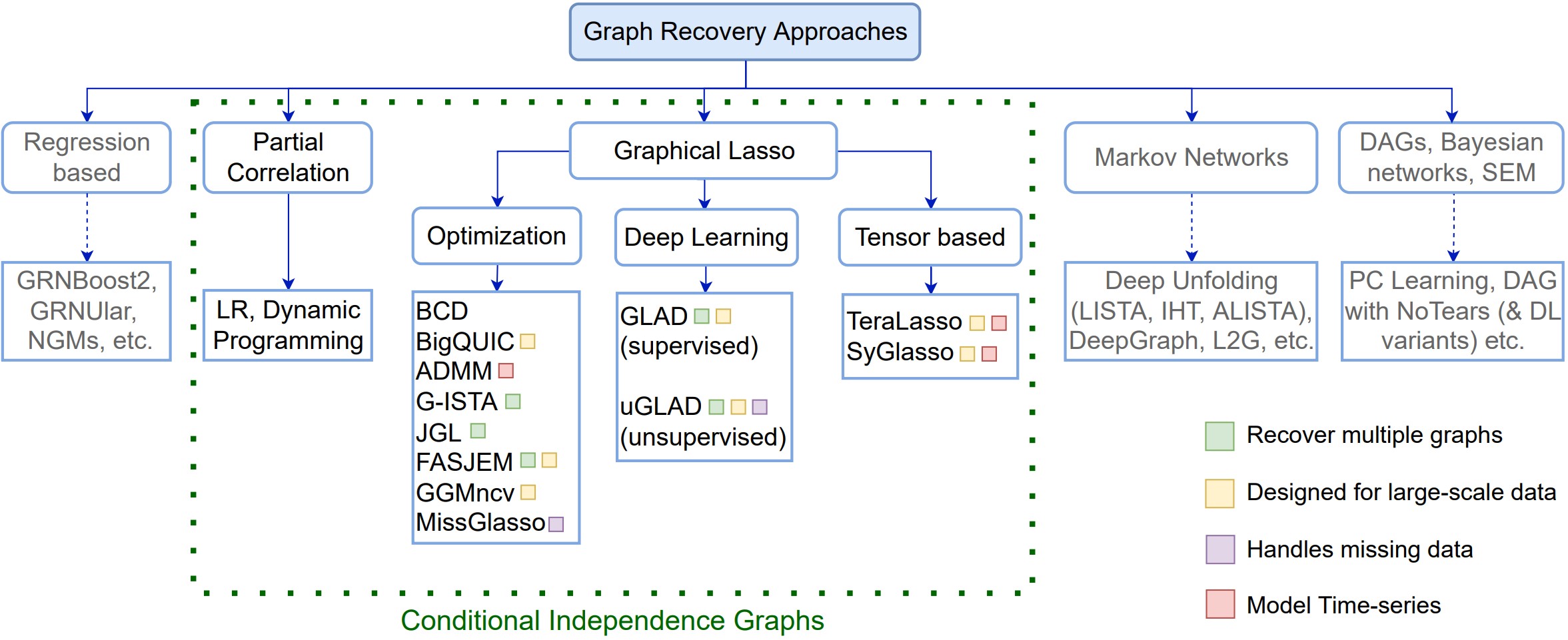}
\caption{\small \textbf{Graph Recovery approaches.} Methods used to recover Conditional Independence graphs are the focus of this survey. The recovered CI graph shows partial correlations between the feature nodes. The algorithms (leaf nodes) listed here are representative of the sub-category and the list is not exhaustive.}
\label{fig:ci-graph-diagram}
\end{figure}

Fig.~\ref{fig:ci-graph-diagram} attempts to list popular formulations of graph representations and then list representative algorithms to recover the same. The algorithms that recover conditional independence graphs parameterized to represent partial correlations are the focus of this survey and are discussed in Sec.~\ref{sec:methods}. For the sake of better understanding of this space, we will briefly describe approaches that recover graphs, edges of which not necessarily represent partial correlations between nodes.

\textit{Regression based.} This line of research follows the idea of fitting regression between features of the input data $\textbf{X}$ to find dependencies among them. This approach is particularly popular for recovering Gene Regulatory Networks (GRN) where the input gene expression data have $D$ genes and $M$ samples, $X\in \mathbb{R}^{M\times D}$. Generally, the objective function used for graph recovery is a variant of the regression between the expression value of each gene as a function of the other genes (or alternatively transcription factors) and some random noise $X_d = f_d\left(X_{D\backslash{d}}\right) +\epsilon$, $\forall{d}\in D$. Usually, a sparsity constraint is also associated with the regression function to identify the top influencing genes for every gene. Many methods have been developed specifically for GRN recovery with varied choice of the regression function $f_d$. TIGRESS~\citep{haury2012tigress} modeled $f_d$ as linear, GENIE3~\citep{van2010inferring} took each $f_g$ to be a random forest while GRNBoost2~\citep{moerman2019grnboost2} used gradient boosting technique. Then, neural network based representations like GRNUlar~\citep{shrivastava2020grnular,shrivastava2022grnular} were developed. Recently proposed Neural Graphical Models (NGMs) ~\citep{shrivastava2022neural}
used neural network as a multitask learning to fit regressions and recover graph for generic input datatypes.

\textit{Markov Networks.} Markov networks are probabilistic graphical models defined on undirected graphs that follow the Markov properties~\citep{koller2009probabilistic}. While Markov networks follow the conditional independence properties (pairwise, local and global), we made the distinction from the CI graphs based on the interpretation of the edge connections. There are traditional constraint-based and score-based structure learning methods to learn Markov networks~\cite{koller2009probabilistic}. Though, these methods suffer from combinatorial explosion of computation requirements and often simplifying approximations are made. Recently,~\cite{belilovsky2017learning} designed a supervised deep learning architecture to learn mapping from the samples to a graph, called `DeepGraph'. They had considerable number of learning parameters, while the performance showed limited success. On the other hand, the `deep unfolding' or `unrolled algorithm' methodology for estimating of sparse vector like Iterative Shrikage Thresholding Algorithm (ISTA)~\cite{gregor2010learning}, ALISTA~\cite{liu2019alista} and others~\citep{sun2016deep,chen2018theoretical,chen2021learning} that were primarily developed for other applications (eg. compressed sensing etc.) have been adopted to recover Markov networks.~\cite{pu2021learning} proposed a deep unfolding approach, named L2G, to learn graph topologies. Their framework can also unroll a primal-dual splitting algorithm~\citep{komodakis2015playing,kalofolias2016learn} into a neural network for their model.

\textit{Directed Graphs.} This is a very active area of research with lots of new methods being developed at a rapid pace. Directed Acyclic Graphs, Bayesian Networks, Structural Equation models are the prominent types of directed graphs of interest. PC-learning algorithm was one of the first techniques developed to learn Bayesian Networks~\cite{spirtes1995learning}, followed by a suite of score-based and constraint-based learning algorithms and their parallel variants~\citep{heckerman1995learning,koller2009probabilistic}. ~\cite{zheng2018dags} introduced `DAG with NOTEARS' method which converted the combinatorial optimization problem of DAG learning to a continuous one. This led to development of many follow up works including some deep learning methods like ~\cite{yu2019dag,zhang2019d,zheng2020learning,pamfil2020dynotears} to name a few.~\cite{heinze2018causal} provide a review of the causal structure learning methods and structural equation models.

\section{Recovering Conditional Independence Graphs}\label{sec:methods}

We define the scope of conditional independence (CI) graphs as undirected probabilistic graphical models that show partial correlations between features. Before we deep dive into the algorithms that recover CI graphs, covered by the green envelope in Fig.~\ref{fig:ci-graph-diagram}, we will provide a useful primer on handling input data with mixed datatypes. Encountering a mix of numerical (real, ordinal) and categorical variables in the input is very common. Since many of the CI graph recovery models take covariance matrix $\operatorname{cov}(X)\in\mathbb{R}^{D\times D}$ as in input, one way to accommodate for categories is to calculate the covariance matrix with the categorical variables included. 

\subsection{Covariance matrix for mixed datatypes}\label{sec:cov-mixed-types}

We describe ways to calculate the covariance matrix for inputs $\textbf{X}$ with $M$ samples and $D$ features consisting of numerical and categorical types. The value of each entry of $\operatorname{cov}({\bf X})\in \mathbb{R}^{D\times D}$ depends on the type of interacting features, say $X_i, X_j$, and will be one of the following:

(I) \textit{numerical-numerical correlation.} The Pearson correlation coefficient, range between $[-1,1]$, defined as $\rho_{X_i, X_j}=\frac{\mathop{{}\mathbb{E}}\left[(X_i-\mu_{X_i})(X_j-\mu_{X_j})\right]}{\sigma_{X_i}\sigma_{X_j}}$, where $\sigma_{X_i}$ denotes standard deviation and $\mu_{X_i}$ is the mean of the feature $X_i$ and similarly for $X_{j}$. The Pearson correlation assumes that the variables are linearly related and can be quite sensitive to outliers. To capture non-linear relationships, ordinal association (or rank correlation) based metrics like Spearman Correlation, Kendall's Tau, Googman Kruskal's Gamma or Somer's D can be leveraged.

(II) \textit{categorial-categorial association.} Cram\'ers V statistic along with the bias correction. The association value is in the range of $[0,1]$, where 0 means no association and 1 is full association between categorical features. Consider two categorical features $C^1$ \& $C^2$ with sample size $M$ and have $i=\{1,\cdots, p\}$, $j=\{1,\cdots, q\}$ possible categories, respectively. Let $m_{ij}$ denote the number of times the values $(C^1_i, C^2_j)$ occur. The Cram\'ers V statistic is defined as $V=\sqrt{\frac{\chi^2/m}{\min{(p-1, q-1)}}}$, where $\chi^2=\sum_{ij}\frac{\left(m_{ij}-\frac{m_{i*}m_{*j}}{m}\right)^2}{\frac{m_{i*}m_{*j}}{m}}$ is the chi-square statistic, $m_{i*}=\sum_{j}m_{ij}$ and $m_{*j}=\sum_{i}m_{ij}$. It has been observed that Cram\'ers V statistic tends to overestimate the association strength. To address this issue, a bias correction modification was introduced as $\Tilde{V}=\sqrt{\frac{\Tilde{\varphi}^2}{\min{(\Tilde{p}-1, \Tilde{q}-1)}}}$, where $\Tilde{\varphi}^2=\max{\left(0, \chi^2/m-\frac{(p-1)(q-1)}{(m-1)}\right)}$, $\Tilde{p}=p-\frac{(p-1)^2}{(m-1)}$ and $\Tilde{q}=q-\frac{(q-1)^2}{(m-1)}$. Some more interesting approaches using `Gini-index' and `word2vec' representation are discussed in~\cite{niitsuma2016word2vec} and can also be utilized. 

(III) \textit{categorical-numerical correlation.} There are multiple options available like the correlation ratio (range is $[0, 1]$)
, point biserial correlation   (range is $[-1, 1]$)
, Kruskal-Wallis test by ranks (or H test)
. Each of these methods have their plus points and drawbacks that should be considered while selecting the metric. Yet another way can be to bin the numerical variable and convert it to a categorical variable. Then Cram\'ers V can be used to calculate the correlation. 

\subsection{Methods}

Recovery of conditional independence graphs is a topic of wide research interest. There are two popular formulations to recover CI graphs, either directly determining partial correlation values or deriving them by utilizing matrix inversion approaches. Based on the formulation chosen, many optimization algorithms have been developed with each having their own capabilities and limitations. 

\subsubsection{Directly calculate partial correlations}


The definition of the partial correlation between two features $X_i$ and $X_j$ given a set of $D-2$ controlling variables $\textbf{X}_{D\backslash{i,j}}$, written $\rho_{X_i, X_j}.\textbf{X}_{D\backslash{i,j}}$, is the correlation between the residuals $e_{X_i}$ and $e_{X_j}$ after fitting a linear regression of $X_i$ with  $\textbf{X}_{D\backslash{i,j}}$ and of $X_j$ with  $\textbf{X}_{D\backslash{i,j}}$, respectively. Poplular approaches used to obtain the partial correlation values directly are discussed below.

    \textit{Linear regression.} The regressions for the partial correlation calculations are formulated using linear functions. Let vectors $\{\textbf{w}_i, \textbf{w}_j\}\in\mathbb{R}^{D-1}$ and $\textbf{X}_{D\backslash{i,j}}$ denote the vector of the other features augmented by $1$ to account for bias. Then the regression over the $M$ samples will be
    \begin{align}
        \textbf{w}_k= \argmin_\textbf{w}{\sum_{m=1}^M X_k^m - \langle \textbf{w}, \textbf{X}_{D\backslash{i,j}}^m \rangle},\quad\text{where k=\{i, j\}.}
    \end{align}
    We then calculate the residuals for each individual sample as $e_{X_k}^m = X_k^m - \langle \textbf{w}, \textbf{X}_{D\backslash{i,j}}^m \rangle$ where k=\{i, j\}. Now, the partial correlation between $X_i, X_j$, which is the $\{i, j\}$ entry of the matrix $\Rho\in\mathbb{R}^{D\times D}$ is calculated as the correlation between the residuals, 
    \begin{align}
        \rho_{X_i, X_j}.\textbf{X}_{D\backslash{i,j}} = \frac{M\sum_{m=1}^M e_{X_i}^m e_{X_j}^m}{\sqrt{M\sum_{m=1}^M (e_{X_i}^m)^2}\sqrt{M\sum_{m=1}^M (e_{X_j}^m)^2}}
    \end{align}
    
    \textit{Recursive formulation.} One major drawback of the linear formulation is it being computationally expensive. The recursive formulation uses dynamic programming based algorithm to recursively calculate the following partial correlation expression, for any $X_k\in \textbf{X}_{D\backslash{i,j}}$,
    \begin{align}
        \rho_{X_i, X_j}.\textbf{X}_{D\backslash{i,j}} = \frac{ \rho_{X_i, X_j}.\textbf{X}_{D\backslash{i,j,k}} -  \rho_{X_i, X_k}.\textbf{X}_{D\backslash{i,j,k}} \times \rho_{X_k, X_j}.\textbf{X}_{D\backslash{i,j,k}}  }{\sqrt{1- \rho^2_{X_i, X_k}.\textbf{X}_{D\backslash{i,j,k}} } \sqrt{1- \rho^2_{X_k, X_j}.\textbf{X}_{D\backslash{i,j,k}}}}
    \end{align}

\subsubsection{Graphical Lasso \& variants}

Given $m$ observations of a $d$-dimensional multivariate Gaussian random variable $X=[X_1,\ldots,X_d]^\top$, the sparse graph recovery problem aims to estimate its covariance matrix $\Sigma^*$ and precision matrix $\Theta^* = (\Sigma^*)^{-1}$. 
The $ij$-th component of $\Theta^*$ is zero if and only if $X_i$ and $X_j$ are conditionally independent given the other variables $\{X_k\}_{\raisemath{1.5pt}{k\neq i,j}}$. The general form of the graphical lasso optimization to estimate $\Theta^*$ is the log-likelihood of a multivariate Gaussian with regularization as
\begin{align}
 \label{eq:glasso-general}
 \widehat{\Theta} =  \argmin\nolimits_{\Theta \in \gS_{++}^d} ~~- \log(\det{\Theta}) +  \text{tr}(\widehat{\Sigma}\Theta) + \operatorname{Reg}(\Theta_\text{off}),
 \end{align}
where $\widehat{\Sigma}$ is the empirical covariance matrix based on $m$ samples, $\gS_{++}^d$ is the space of $d\times d$ symmetric positive definite matrices and $\operatorname{Reg}(\Theta_\text{off})$ is the regularization term for the off-diagonal elements. Once the precision matrix is obtained, the corresponding partial correlation matrix entries can be calculated as 
$\rho_{X_i, X_j}.\textbf{X}_{D\backslash{i,j}} =-\frac{\widehat{\Theta}_{i,j}}{\sqrt{\widehat{\Theta}_{i,i} \widehat{\Theta}_{j,j}}}$. Several algorithms have been developed to optimize the sparse precision matrix estimation problem in Eq.~\ref{eq:glasso-general} which primarily differ in the choice of regularization and optimization procedure. 

\textit{Block Coordinate Descent (BCD).} 
~\cite{banerjee2008model} formulated the graphical lasso problem of approximating precision matrix as the $\ell_1$-regularized maximum likelihood estimation
\begin{align}
 \label{eq:sparse_concentration-theta}
 \widehat{\Theta} =  \argmin\nolimits_{\Theta \in \gS_{++}^d} ~~- \log(\det{\Theta}) +  \text{tr}(\widehat{\Sigma}\Theta) + \lambda\norm{\Theta}_{1,\text{off}},
 \vspace{-1mm}
 \end{align}
 where $\norm{\Theta}_{1,\text{off}}=\sum_{i\neq j}|\Theta_{ij}|$ is the off-diagonal $\ell_1$ regularizer with regularization parameter $\lambda$.  Block-coordinate Descent  methods, for example ~\citep{friedman2008sparse}, updates each row (and the
corresponding column) of the precision matrix iteratively by solving a sequence of lasso problems. A variant of this algorithm is the popular `GraphicalLasso' implementation of python's scikit-learn package~\cite{scikit-learn} and is very efficient for large scale problems involving thousands of variables. This estimator is sensible even for non-Gaussian input data, since it is minimizing an $\ell_1$-penalized log-determinant Bregman divergence~\citep{ravikumar2011high}. Different variants of solvers have been proposed to handle large scale data, some of the prominent ones being \texttt{BigQUIC, QUIC, SQUIC} by ~\cite{hsieh2013big,hsieh2014quic,bollhofer2019large} respectively, can handle up to 1 million random variables.


\textit{G-ISTA.} The Graphical Iterative Shrinkage Thresholding Algorithm was proposed by \cite{rolfs2012iterative}. This method uses proximal gradient descent based approach to perform $\ell_1$-regularized inverse covariance matrix estimation specified in Eq.~\ref{eq:sparse_concentration-theta}. The basic idea is to separate the continously differentiable, convex function (first two terms of Eq.~\ref{eq:sparse_concentration-theta}) and the regularization term which is convex but not necessarily smooth. Then the standard updates of general iterative shrinkage thresholding algorithm (ISTA)~\citep{beck2009fast} can be applied. The G-ISTA algorithm comes with nice theoretical and stability properties. Similarly, Alternating Direction Method of Multipliers (ADMM)~\citep{boyd2011distributed} can be used to optimize the $\ell_1$ regularized objective.

\begin{wrapfigure}[33]{R}{0.21\textwidth}
\vspace{-2mm}
\centering
\includegraphics[width=25mm, height=105mm]{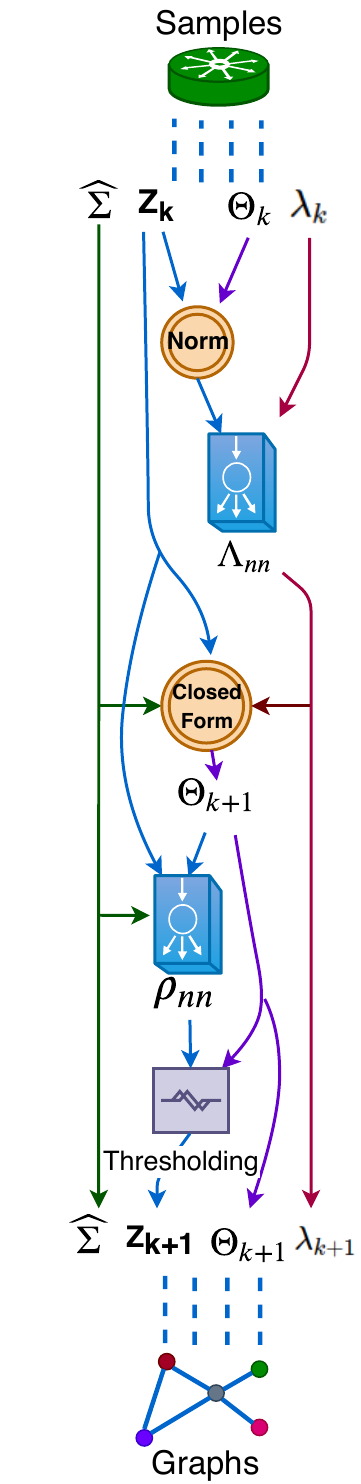}
\caption{\small The recurrent unit \texttt{GLADcell}. (Taken from~\cite{shrivastava2020glad})} 
\label{fig:glad-architecture}
\end{wrapfigure}

\textit{Graphical non-convex optimization.}~\citep{pmlr-v80-sun18c} proposed graphical non-convex optimization for optimal estimation in Gaussian Graphical models. They consider the optimization objective  of Eq.~\ref{eq:glasso-general} with the $\operatorname{Reg}(\Theta_\text{off})=\sum_{i\neq j}\lambda(\Theta_{i,j})$, where $\lambda(\cdot)$ is a non-convex penalty. This non-convex optimization is then approximated by a sequence of adaptive convex programs. The experiments demonstrate improvements over previous methods and thus advocate for development of methods that account for non-convex penalties. They note that their algorithm is an adaptive version of the SPICE algorithm by~\cite{rothman2008sparse}. Another recent work 
by~\cite{zhang2018large} builds up on prior work that shows the the graphical lasso estimator can be retrieved by soft-thresholding the sample covariance matrix and solving a maximum determinant matrix completion (MDMC) problem. Their method proposed a Newton-CG algorithm to efficiently
solve the MDMC problem and the authors claim it to be highly efficient for \textbf{large scale} data. We note that there are many works which model sparsity constraints differently or provide surrogate objective functions. We list a few prominent ones here to help the readers get an overview~\citep{loh2011high,yang2014elementary,sojoudi2016equivalence,zhang2018large}. A collection of methods, \texttt{GGMncv}, on Gaussian Graphical models with non-convex regularization can be found in ~\citep{williams2020beyond}.

\textit{\texttt{GLAD}.}~\cite{shrivastava2019glad,shrivastava2020using} proposed a supervised deep learning based model, presented in Fig.~\ref{fig:glad-architecture} to recover sparse graphs based on the graphical lasso objective. They built up on the theoretically proven advantages of having non-convex penalty in the graphical lasso objective. They also proved that it is beneficial to have adaptive sequence of penalty hyperparameters for the regularization term as it leads to faster convergence. Specifically, they applied the Alternating Minimization (AM) algorithm to the objective in Eq.~\ref{eq:sparse_concentration-theta} and unrolled the AM algorithm to certain number of iterations, also known as `deep unfolding' technique. Then, the hyperparamters were parameterized using small neural networks for doing operations like entry-wise thresholding of the precision matrix. Their deep model \glad has significantly fewer number of learnable parameters along with being fully interpretable as one can inspect the recovered graph at any point of optimization. \glad was also successful in avoiding any post-processing requirements to maintain the symmetric and SPD properties of the recovered precision matrix. The training of \glad model was done using supervision and the hope was that the model can generalize over that underlying distribution of graphs. They were first to demonstrate that learning can also help improve sample complexity.

\textit{\texttt{uGLAD}.} Proposed by~\citep{shrivastava2022a,shrivastava2022uglad}, \uglad is an unsupervised deep model that circumvents the need of supervision which was the key bottleneck of the \glad model. They changed the optimization problem by introducing the `glasso' loss function and incorporating the regularization in the deep model architecture (defined by \glad) itself which is implictly learnt during optimization. The input of \uglad is the empirical covariance matrix; it requires no ground-truth information to optimize. It can also do multitask learning by optimizing \textbf{multiple CI graphs} at once. Some other prominent methods that recover multiple graphs include \texttt{JGL}~\citep{danaher2014joint} and \texttt{FASJEM}~\citep{wang2017fast}. Furthermore, \uglad leverages this ability to handle \textbf{missing data} by introducing a consensus strategy and thus have more robust performance. A different approach to handle missing data have also been previously explored in \texttt{MissGLasso}~\citep{stadler2012missing}, ~\citep{loh2011high} among other methods.

\textit{\texttt{Tensor graphical lasso}.}   ~\cite{greenewald2019tensor} proposed \texttt{TeraLasso} that extends the graphical lasso problem to higher-order tensors. They introduced a multiway tensor generalization of the bi-graphical lasso which uses a two-way sparse Kronecker sum multivariate normal model for the precision matrix to model parsimoniously conditional dependence relationships of matrix variate data based on the Cartesian product of graphs. The Sylvester Graphical Lasso or \texttt{SyGlasso} model by ~\cite{wang2020sylvester} complements \texttt{TeraLasso} by providing an alternative Kronecker sum model that is generative and interpretable. These approaches are typically very helpful in modeling spatio-temporal data. 

\begin{table}[]
\centering
\tiny
\begin{tabular}{|c|c|c|c|c|c|c|}
\hline
\rowcolor[HTML]{CBCEFB} 
Methods & Implementation & {MG} & {LS} & {MI} & {TS} & {Paper} \\ \hline
\texttt{BCD} & \href{https://scikit-learn.org/stable/modules/generated/sklearn.covariance.GraphicalLassoCV.html}{\texttt{Scikit-learn package}} &  &  &  &  &\cite{scikit-learn}\\ \hline
\texttt{QUIC,BigQUIC} &\href{https://bigdata.oden.utexas.edu/software/1035/}{\texttt{Software link}}  &  & \checkmark &  &  &\cite{hsieh2013big,hsieh2014quic}\\ \hline
\texttt{ADMM} &\url{https://github.com/tpetaja1/tvgl} &  &  &  & \checkmark &\cite{hallac2017network}\\ \hline
\texttt{G-ISTA} & \href{https://github.com/Harshs27/GLAD/tree/master/glad_module/direct}{\texttt{Python package}} &\checkmark  &  &  &  &\cite{rolfs2012iterative}\\ \hline
\texttt{JGL} & \href{https://cran.r-project.org/web/packages/JGL/index.html}{\texttt{R package}} &\checkmark  &  &  &  &\cite{danaher2014joint}\\ \hline
\texttt{FASJEM} & \url{https://github.com/QData/FASJEM} &\checkmark  & \checkmark &  &  &\cite{wang2017fast}\\ \hline
\texttt{GGMncv} & \href{https://cran.r-project.org/web/packages/GGMncv/readme/README.html}{\texttt{R package}} & & \checkmark  &  &  &\cite{williams2020beyond}\\ \hline
\texttt{Newton-CG(MDMC)} & \href{https://ryz.ece.illinois.edu/software.html}{\texttt{Matlab package}} &  &\checkmark  &  &  &\cite{zhang2018large}\\ \hline
\texttt{MissGlasso} & \href{https://www.rdocumentation.org/packages/cglasso/versions/1.1.2/topics/mglasso}{\texttt{R package}} & &  & \checkmark  &  &\cite{stadler2012missing}\\ \hline
\texttt{TeraLasso} & \url{https://github.com/kgreenewald/teralasso} & & \checkmark &  &\checkmark    &\cite{greenewald2019tensor}\\ \hline
\texttt{SyGlasso}& \url{https://github.com/ywa136/syglasso} &  &\checkmark  &  &\checkmark  &\cite{wang2020sylvester}\\ \hline
\glad & \url{https://github.com/Harshs27/GLAD}&\checkmark  &\checkmark  &  &  &\cite{shrivastava2019glad}\\ \hline
\uglad & \url{https://github.com/Harshs27/uGLAD} &\checkmark  & \checkmark  &\checkmark   &  &\cite{shrivastava2022uglad}\\ \hline
\end{tabular}
\caption{Conditional Independence graph recovery methods with their implementation links. Additional information about their ability to recover multiple graphs (MG), handle large scale data (LS), handle missing values in data (MI) and to model time-series (TS) are mentioned alongside.}
\label{tab:methods-ci-graphs}
\end{table}


Table~\ref{tab:methods-ci-graphs} lists some of the prominent methods for CI graph recovery along with their recommended implementations. This compilation will help the readers choose the right models for their applications. Now that we have discussed some of the popular approaches to recover CI graphs, for the sake of completeness and wider adoption, we list out some of the potential applications that the CI graphs have been applied to in the past as well as hint at several unexplored opportunities to leverage them. 

\section{Potential applications of CI graphs}\label{apx:applications}

Listing some potential applications where the CI graph recovery algorithms can be applied with a potential of improvement over the current state-of-the-art.

\begin{figure}
    \centering
    \includegraphics[width=0.49\textwidth, height=50mm]{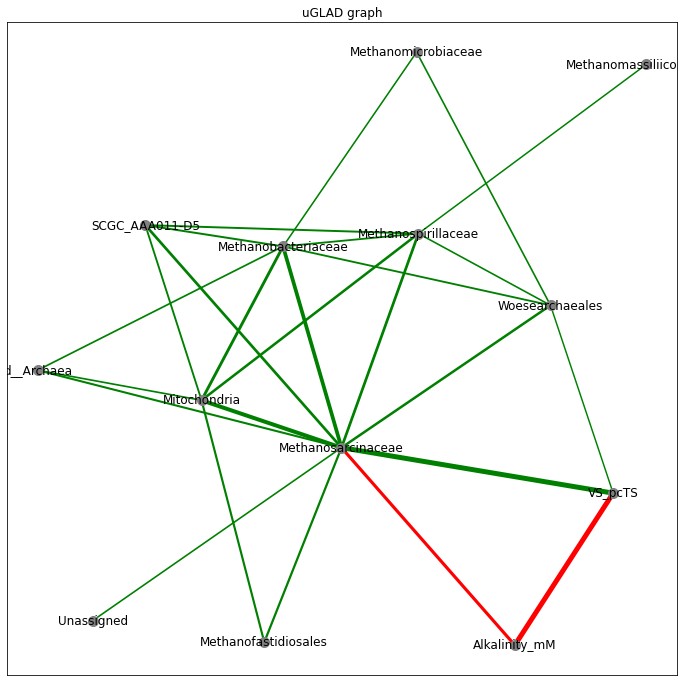}~
    \includegraphics[width=0.49\textwidth, height=50mm]{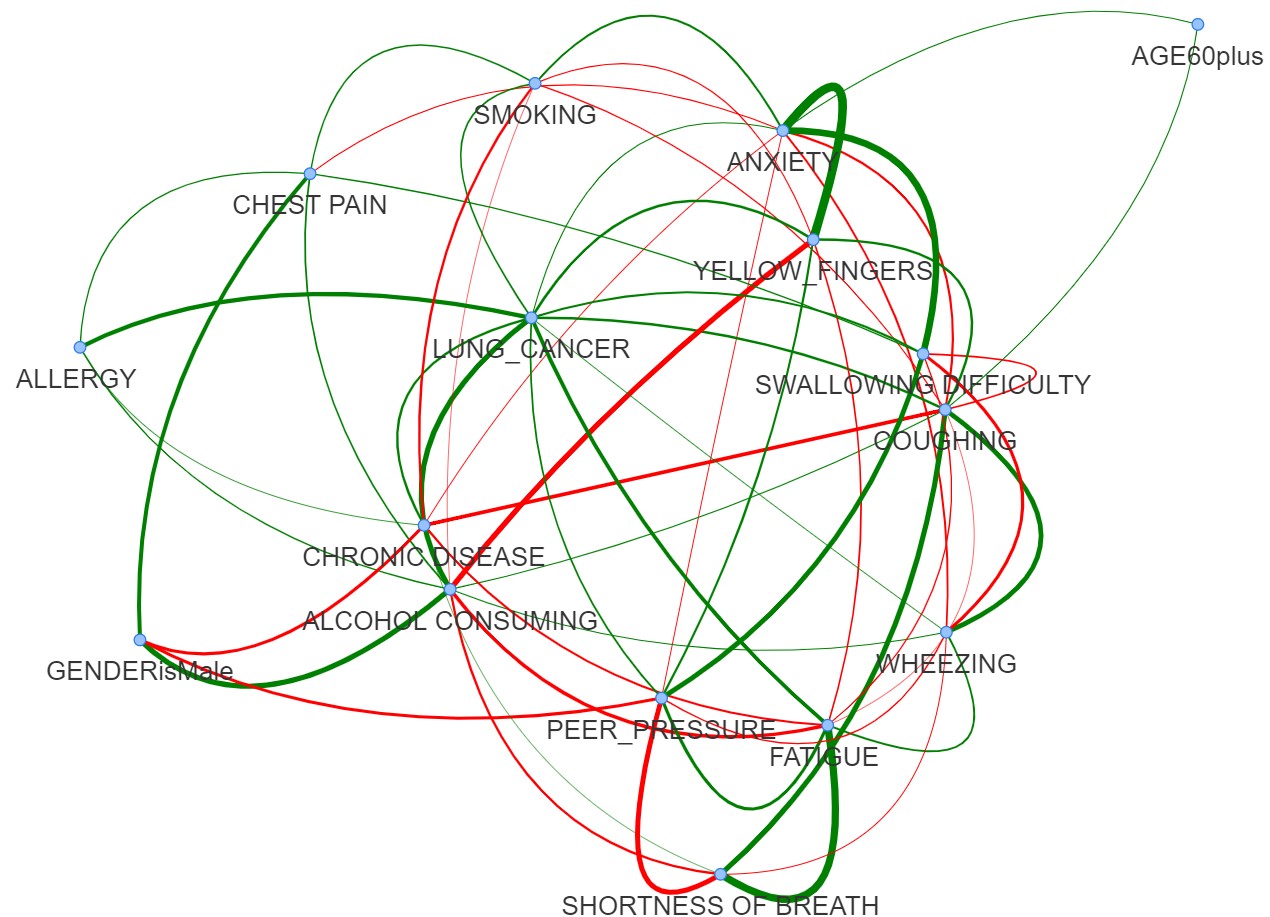}
    \caption{\small [left] \uglad graph for archaea at family level in a collection of wastewater processing digesters. Edge color indicates the sign of the correlation: green - positive, red - negative, edge weight corresponds to correlation's strength (taken from~\citep{shrivastava2022uglad}). [right] CI graphs from \uglad model used to analyse a lung cancer data from~\cite{lcData}.}
    \label{fig:life-sciences-uglad}
\end{figure}

\begin{figure}[h]
\vspace{-2mm}
\subfigure[True graph]{\includegraphics[width=42mm]{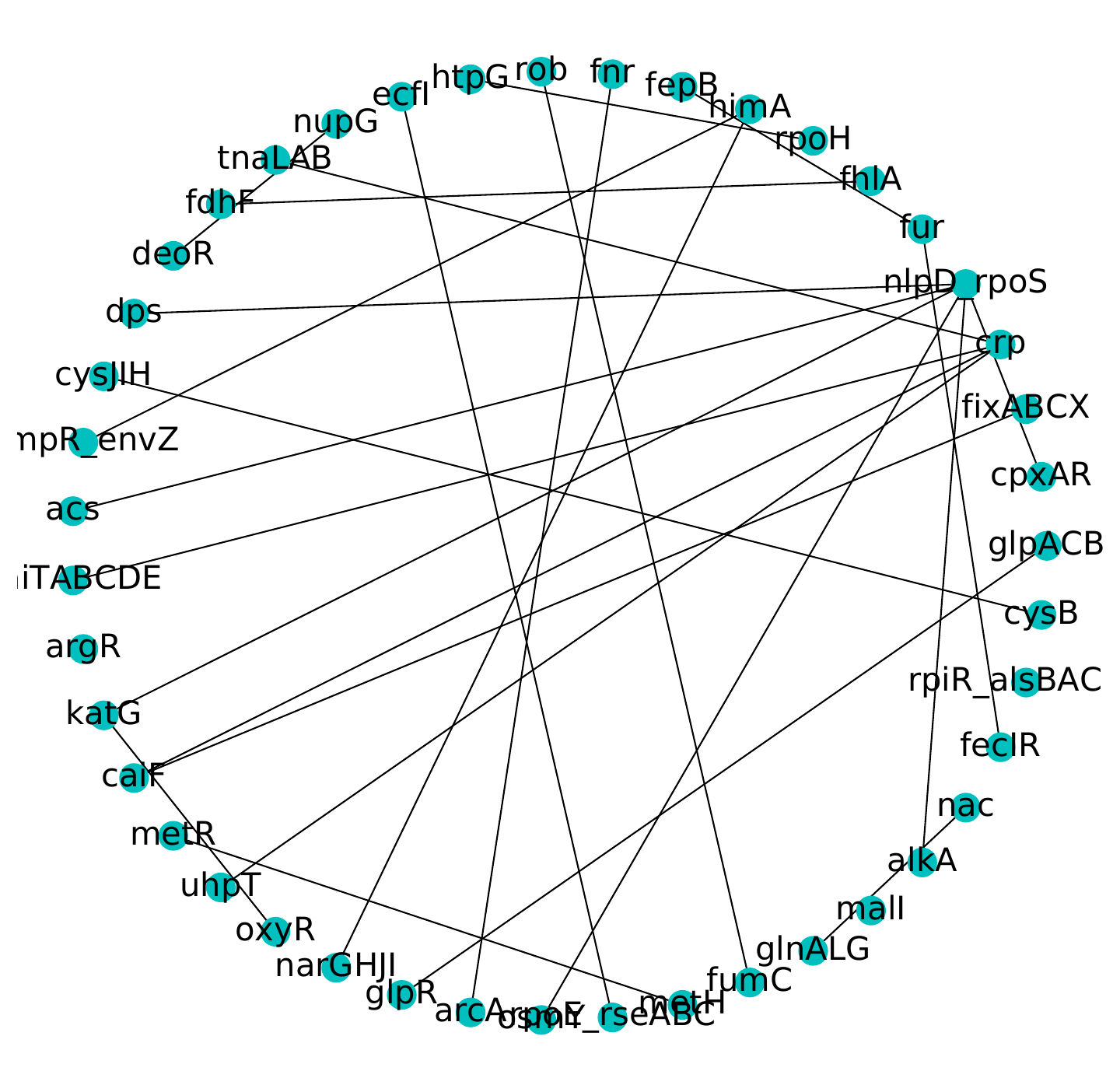}}
\subfigure[\textbf{M=10}, {\color{red}fdr=0.613}, {\color{green!60!black}tpr=0.913}, fpr=0.114 ]{\includegraphics[width=42mm]{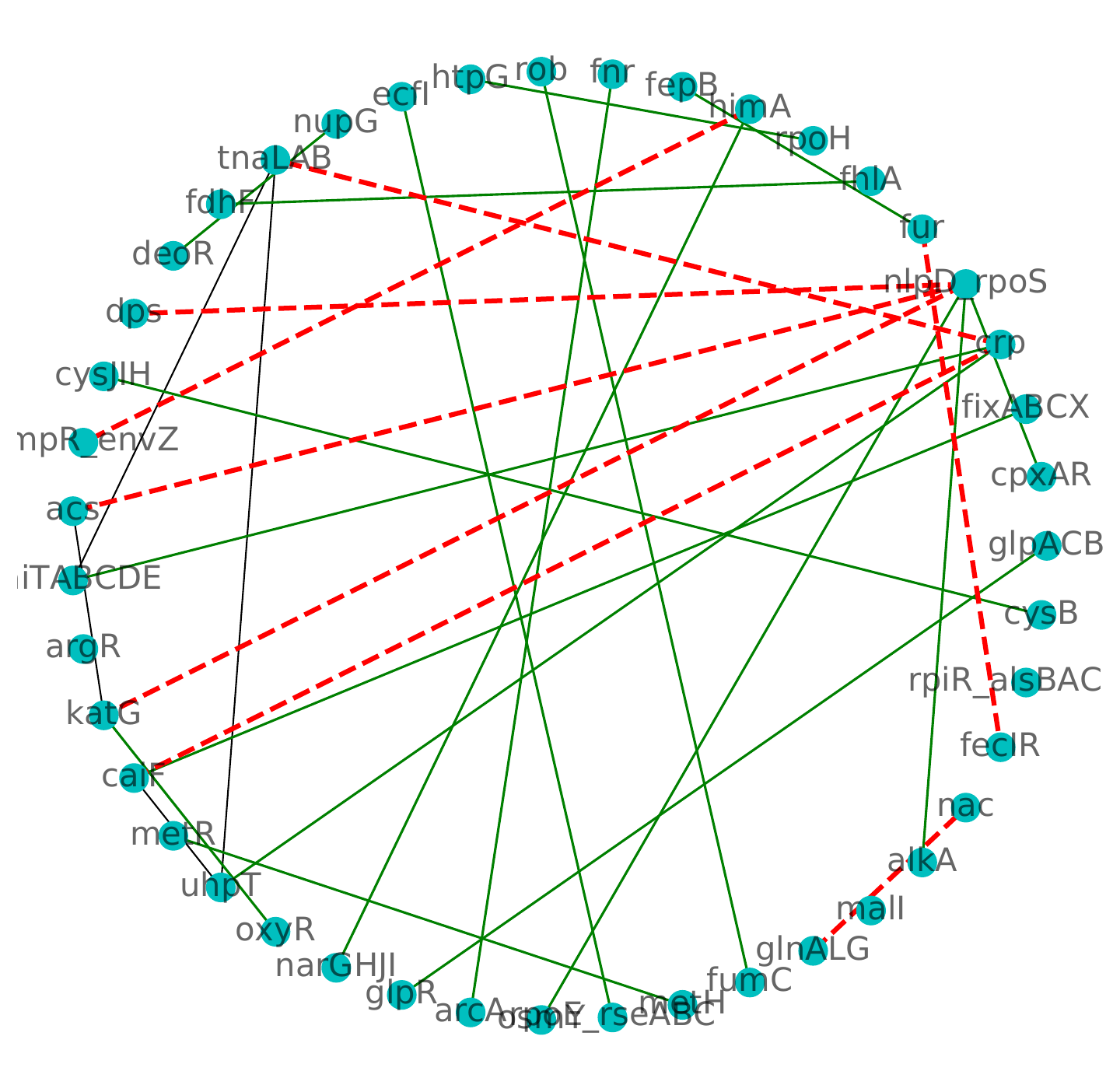}}\hspace{10pt}
\subfigure[\textbf{M=100}, {\color{red}fdr=0.236}, {\color{green!60!black}tpr=0.986}, fpr=0.024]{\includegraphics[width=42mm]{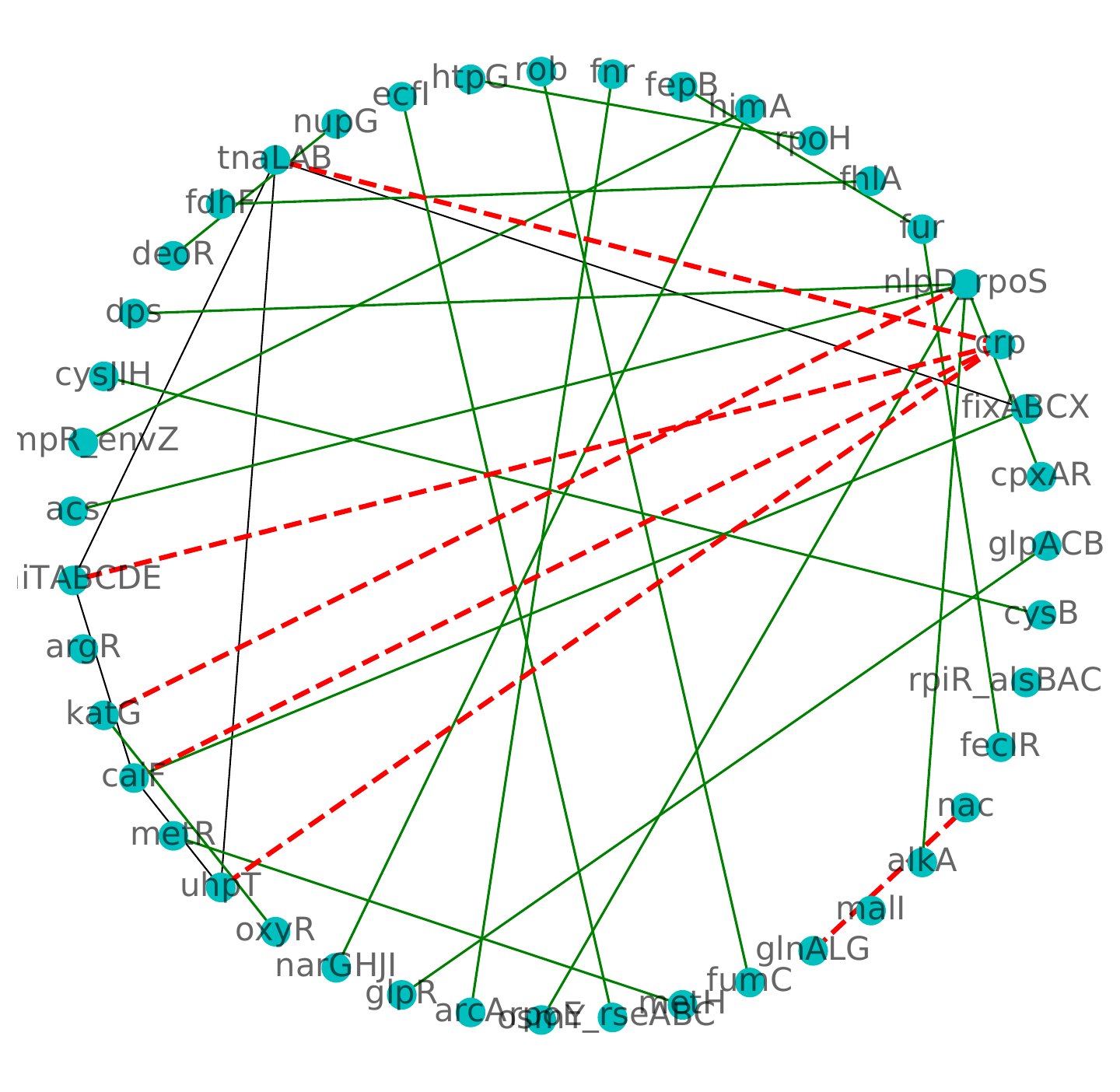}}
{\caption{Recovered graph structures for a sub-network of the {\it E. coli} consisting of $43$ genes and $30$ interactions with increasing samples. \glad was trained using ground truth from a synthetic gene expression data simulator. Increasing the samples reduces the FDR by discovering more true edges.  We denote, TPR: True Positive Rate, FPR: False Positive Rate, FDR: False
Discovery Rate. (taken from~\cite{shrivastava2019glad}). }\label{fig:grn-using-glad}}
\end{figure}

\textit{Life sciences.} CI graphs were successfully used to study the microbials inside an anaerobic digester and to help decide system design parameters of the digester, see Fig.~\ref{fig:life-sciences-uglad}~\cite{shrivastava2022uglad}. Recovering GRNs from the corresponding microarray expression data and possibly extending to ensemble methods~\cite{guo2016gene,aluru2022engrain} can also be interesting to explore using CI graph recovery methods. Recovered GRNs using \glad are shown in Fig.~\ref{fig:grn-using-glad}. CI graphs also naturally align with text mining tools~\cite{roche2017valorcarn,fize2017geodict,antons2020application} and can be leveraged to improved interpretability and performance.

\textit{Class imbalance handling.} The correlations discovered by the CI graphs can be helpful in narrowing down important feature clusters for identifying key features. This will in-turn improve performance in cases where there is little data or imbalanced data (more data points for one class than another). Sampling from these graphs can balance out the data, similar to the SMOTE~\citep{chawla2002smote} procedure. CI graphs can act as preprocessing steps for some of the methods for class imbalance handling like~\cite{rahman2013addressing,shrivastava2015classification,bhattacharya2017icu}.

\textit{Medical Informatics.} Graphical models have been widely used for making informed medical decisions. For instance, the PathFinder project by ~\cite{heckerman1992toward1,heckerman1992toward2} is a prime example of an early system using Bayesian Networks for assisting medical professionals in making critical decisions. The conditional dependencies for this project was procured by consulting the doctors which can be quite time consuming and not easily scalable. Since then, many other medical expert systems based on the Bayesian networks (with parameters usually learned from data) have been developed, see~\cite{MCLACHLAN2020101912} for a survey. CI graphs can be used as an alternative to approximate a distribution over medical variables. For instance, Fig.~\ref{fig:life-sciences-uglad} shows a CI graph for studying patients data for Lung cancer prediction from a Kaggle dataset.

\begin{figure}
\centering 
\includegraphics[width=120mm]{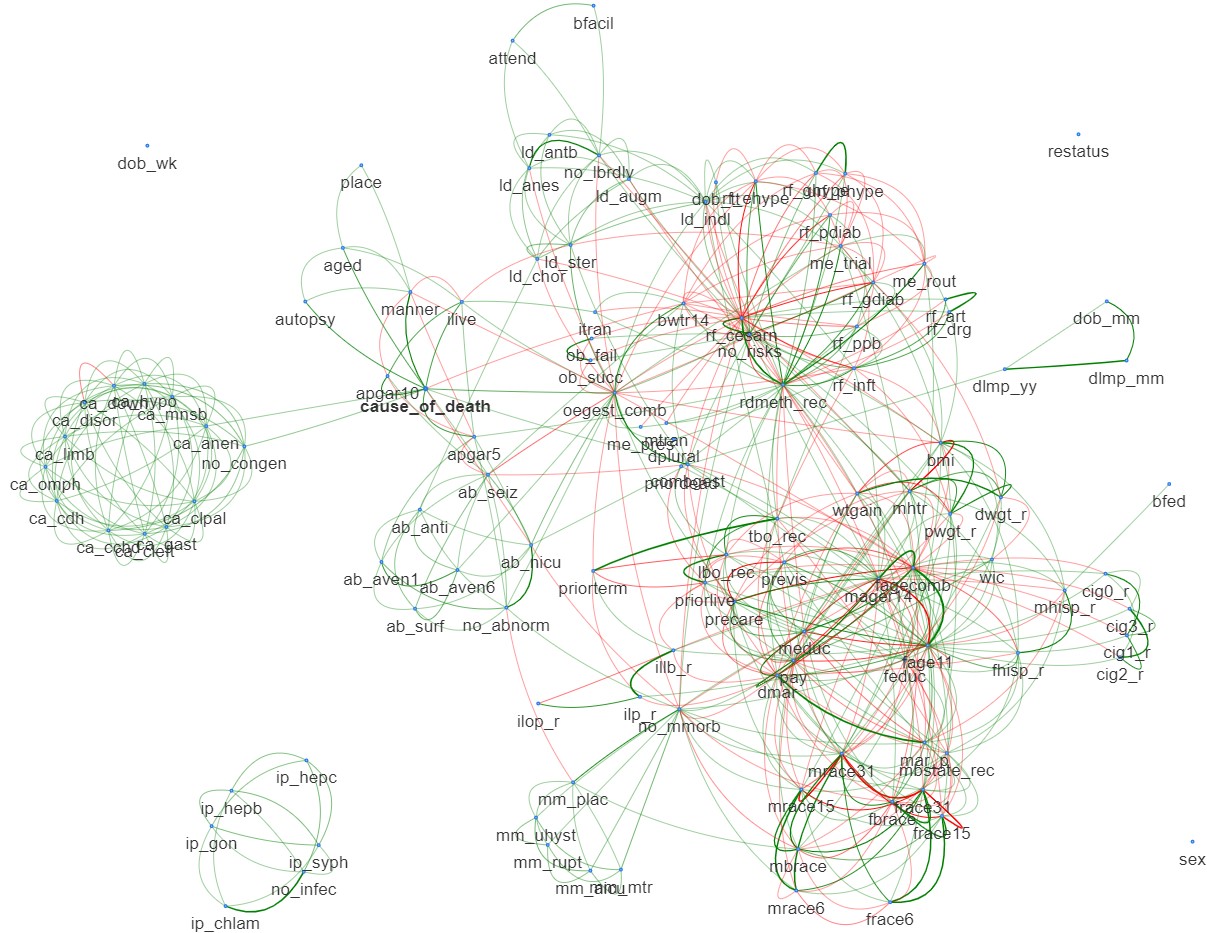}
\caption{\small The CI graph recovered by \uglad for the Infant Mortality 2015 data from CDC~\citep{CDC:InfantLinkedDatasets} (taken from~\cite{shrivastava2022neural}). 
}
\label{fig:infant-mortality-uglad}
\end{figure}

\textit{Finance \& Healthcare.} CI graphs are useful for finding correlations between stocks to see how companies compare~\citep{hallac2017network}. Similarly, CI graphs are the basis of systems for discovering dependencies between important body vitals of ICU patients~\cite{bhattacharya2019methods,shrivastava2021system}.  Another instance is shown in Fig.~\ref{fig:infant-mortality-uglad}, where the authors used a CI graph recovery algorithm to analyse the feature connections to study infant mortality in the US. 

\textit{Video representation \& generation.}  Recent developments in video representation learning can specially benefit from such sparse graph recovery methods. Videos can be interpreted as a sequential collection of frames. Tasks like action recognition within videos~\cite{saini2022recognizing}, prediction of interaction activity~\cite{bodla2021hierarchical}, video segmentation\cite{zhou2022survey} \& object detection\cite{jiao2021new} are of great interest to researchers and have tremendous downstream applications~\cite{oprea2020review}. In some cases, the inter-frame relations can be modeled as temporal processes. Techniques for video prediction by modeling videos as continuous multi-dimensional processes~\cite{shrivastava2024video1}, video priors representation learning methods like~\cite{shrivastava2024video2,shrivastava2024video3} can specially benefit from CI graphs that models similarity across time frames. Such methods can be developed to obtain efficient performance over variety of video tasks and improve model robustness. Deep models for CI graph recovery, \glad or \texttt{uGLAD}, can be integrated into the pipeline for latest models used for generating diverse video frames~\cite{denton2018stochastic,shrivastava2021diverse,shrivastava2021diversethesis}. Specifically, in conjunction with the generative deep models, the CI graph recovery model parameters can be learned to narrow down the potential future viable frames from the generated ones.

\begin{figure}
\centering 
\includegraphics[width=60mm]{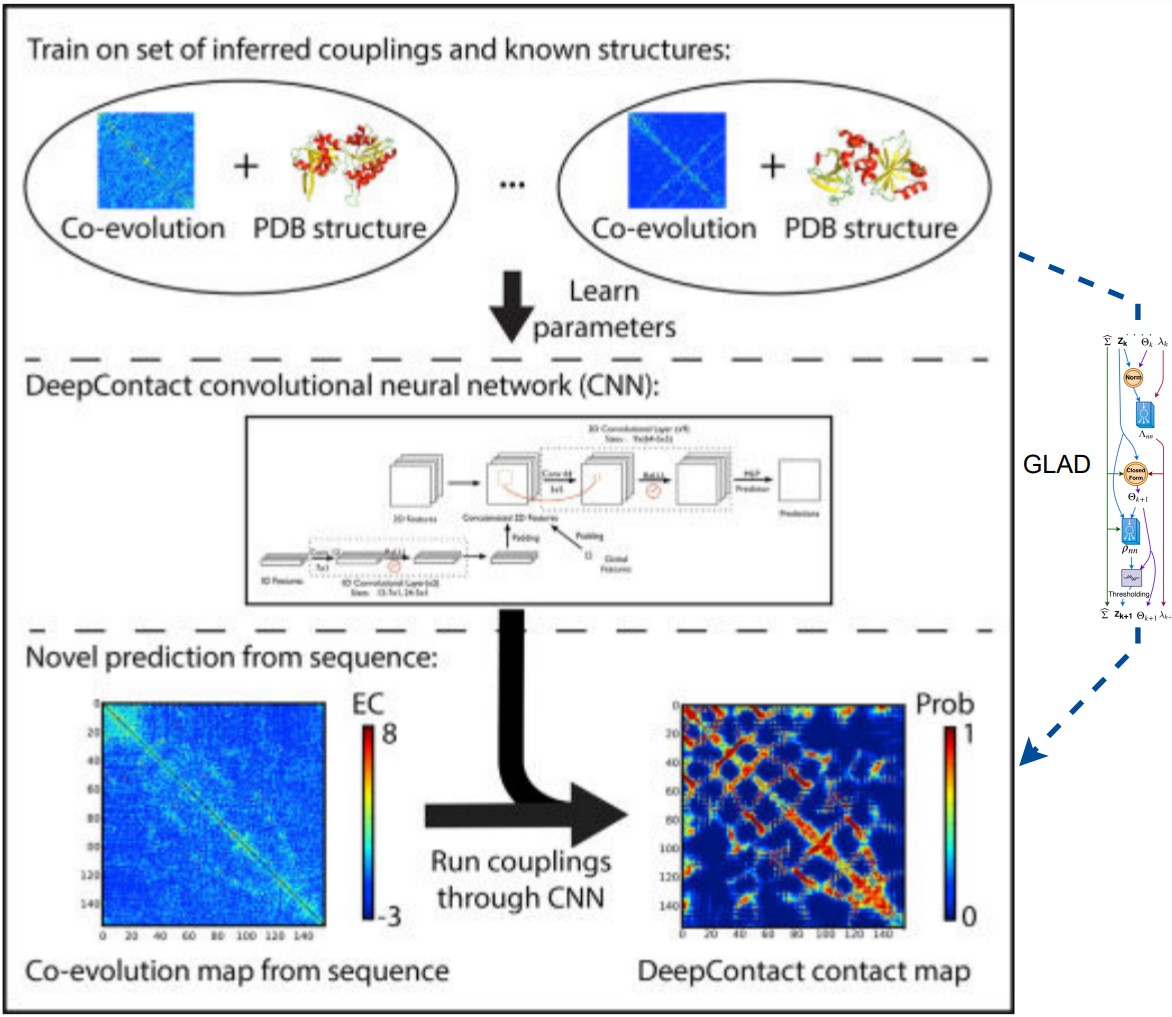}~~~~\qquad
\includegraphics[width=60mm]{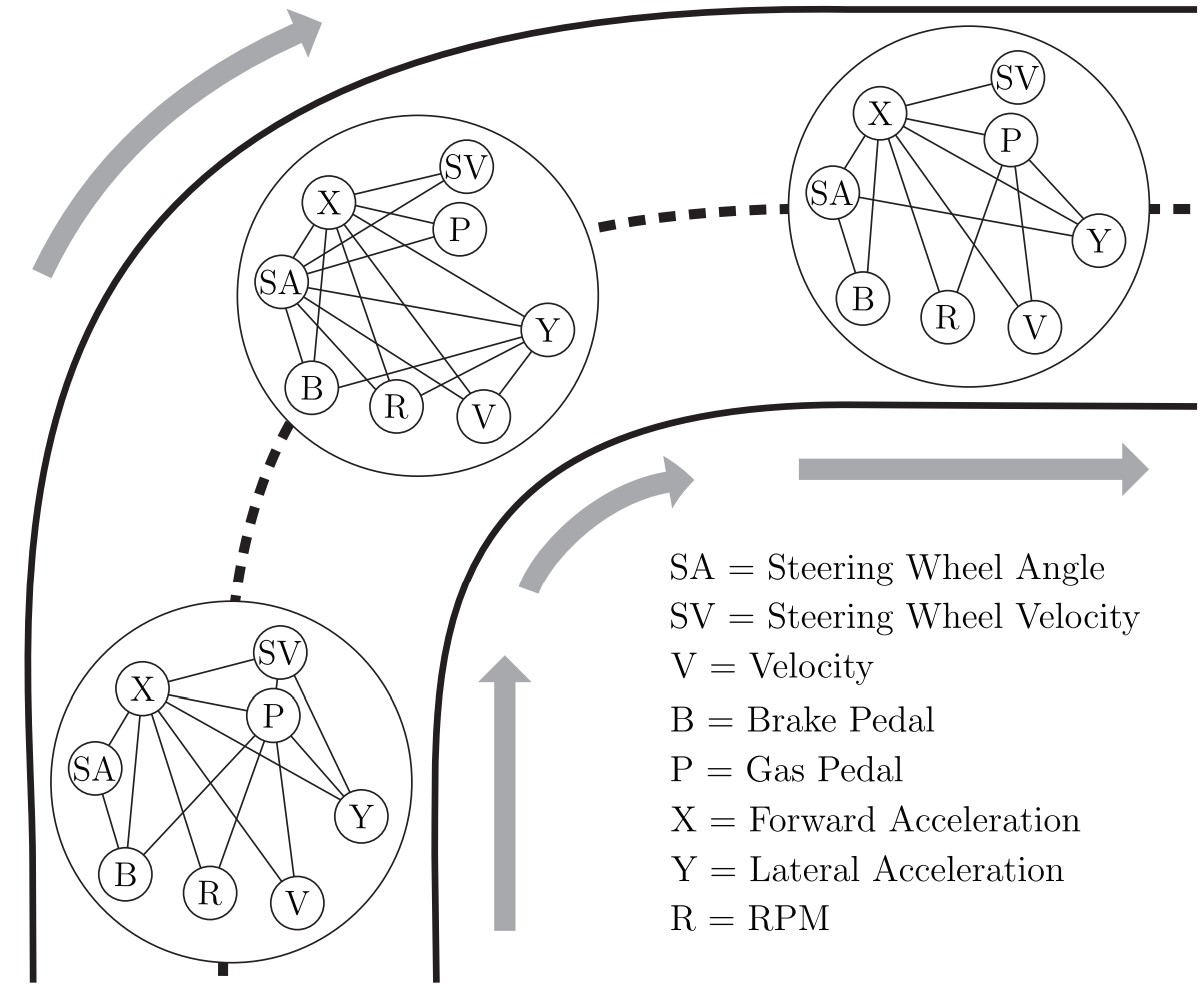}
\caption{\small [left] Potential use of \glad or \texttt{uGLAD} for contact map prediction to recover protein structure. [right] A dynamic network inference framework showing three snapshots of the automobile sensor network
measuring eight sensors, taken (1) before, (2) during, and (3)
after a standard right turn. CI graphs were recovered by running a   scalable message passing algorithm based on the Alternating Direction Method of
Multipliers (ADMM) to study automobile sensor network (taken from~\cite{hallac2017network}). 
}
\label{fig:motor-sensor}
\end{figure}

\begin{figure}
\centering 
\includegraphics[width=120mm]{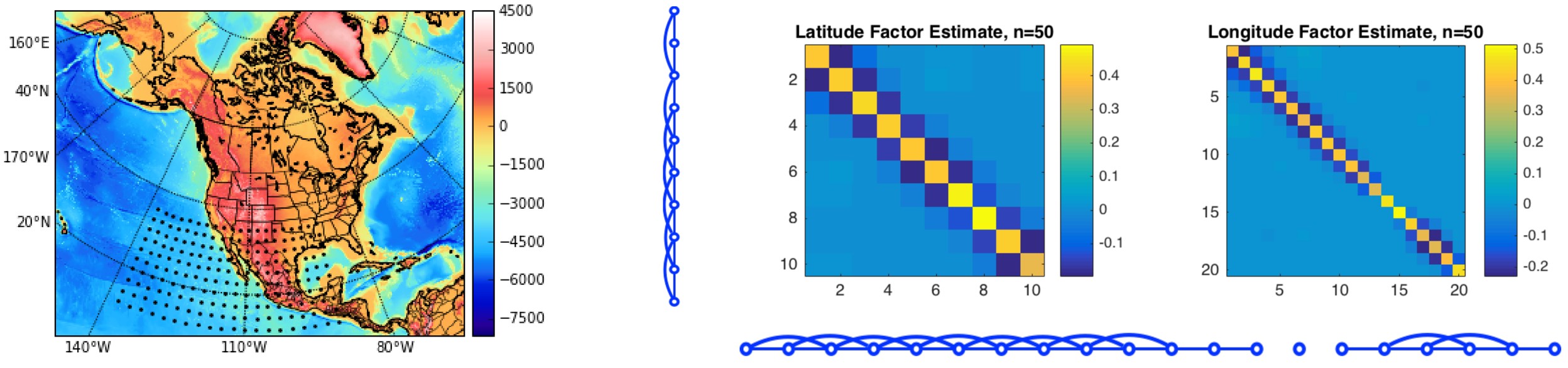}
\caption{\small \texttt{TeraLasso}, a tensor based method, used for analysis of a time series data of daily-average wind speeds of `western grid' of North America. [left] Rectangular 10 × 20 latitude-longitude grids of windspeed locations shown as black dots. Elevation colormap shown in meters. [right] Graphical representation of latitude (left) and longitude factors (bottom) with the corresponding precision estimates. Observe the decorrelation (longitude factor entries connecting nodes 1-13 to nodes 14-20 are
essentially zero) in the Western longitudinal factor, corresponding to the high-elevation line of the Rocky Mountains. (taken from~\cite{greenewald2019tensor}). 
}
\label{fig:weather}
\end{figure}

\begin{figure}
\centering 
\includegraphics[width=120mm]{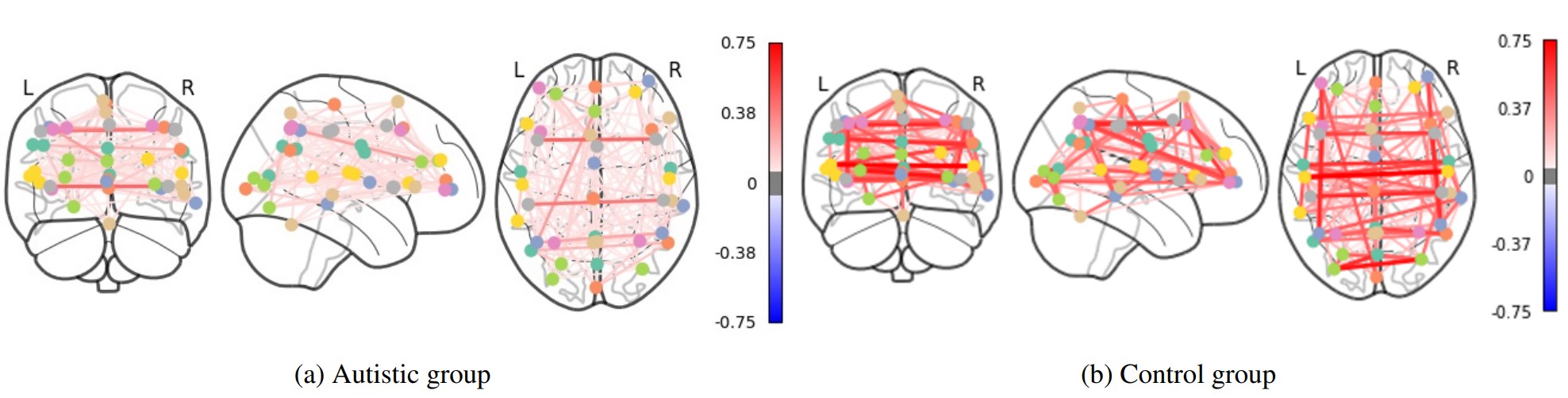}
\caption{\small The connectivity of the 39 regions in the brain estimated by using 35 subjects. The CI graph was recovered by the `L2G' algorithm, which is a deep unfolding approach to learn graph topologies, refer~\cite{pu2021learning}. 
}
\label{fig:autism}
\vspace{-3mm}
\end{figure}

\textit{Protein Structure recovery.} Deep models for CI graph recovery like \uglad can be substituted for predicting the contact matrix from the input correlation matrix between the amino acid sequences. For instance, Protein Sparse Inverse COVaraince or PSICOV~\cite{jones2012psicov},  which uses graphical lasso based approach to predict the contact matrix to eventually predict the 3D protein structure can leverage these recently developed CI graph recovery deep models. Learnable parameters of these models can also account for the ground truth data, if available. DeepContact model~\citep{liu2018enhancing} uses a Convolutional Neural Network based architecture to do a matrix inversion operation for predicting contact map from the co-evolution map obtained from protein sequences. Deep models for CI graph recovery can potentially augment (or even replace) the CNNs for improved predictions, refer to the left side of Fig.~\ref{fig:motor-sensor}.

\textit{Gaussian processes \& time series problems.} An interesting use-case by~\cite{chatrabgoun2021learning} on combining graphical lasso with Gaussian processes for learning gene regulatory networks. Similarly, in a recent work on including negative data points for the Gaussian processes~\cite{shrivastava2020learning}, CI graphs can be used for narrowing down the relevant features for doing the GP regression and for time-series modeling~\citep{jung2015graphical}. An example to understand an automobile system is shown in Fig.~\ref{fig:motor-sensor} on the right~\citep{hallac2017network}.~\cite{greenewald2019tensor} used tensor based formulation of graphical lasso to analyse spatio-temporal data of daily-average wind speeds as shown in Fig.~\ref{fig:weather}. Another interesting analysis was done for inspecting the brain functional connectivity of autism from blood-oxygenation-level-dependent time series as shown in Fig.~\ref{fig:autism} by ~\cite{pu2021learning}. The ability to recover a batch of CI graphs in parallel was utilized for doing multivariate time-series segmentation~\cite{imani2023tglad}. 

\textit{Running Graph Neural Networks over CI graphs.} Various GNN based techniques can be learnt over the CI graph. Especially, methods that are designed to run on Probabilistic Graphical models like the Cooperative Neural Networks~\citep{shrivastava2018cooperative}, Bayesian Deep Learning methods~\citep{wilson2020case} and GNNs developed for other applications~\citep{duvenaud2015convolutional,henaff2015deep,battaglia2018relational} can be potentially adopted. Neural Graphical Models~\citep{shrivastava2022neural} can potentially learn richer distributions compared to feature dependencies discovered in the conditional independence graph. 


\section{Conclusion}

In this survey on the recently developed graph recovery methods, we attempted to build a case for conditional independence graphs for analysis of data from various domains. We provided a breakdown of different methods, traditionally used, as well as recently developed deep learning models, for CI graph recovery along with a primer on their implementation and functioning. In order to facilitate wider adoption, this work also provided various approaches and best practices to handle input data with mixed datatypes which is usually a critical preprocessing step and tricky to manage. We laid out several use-cases of the CI graphs with hope that they will become one of the mainstream methods for data exploration and insight extraction.

\bibliography{bibfile}
\bibliographystyle{iclr2023_conference}


\end{document}